# Human Head Pose Estimation by Facial Features Location

[Eugene Borovikov](#)

University of Maryland Institute for Computer Studies, College Park, MD 20742

4/21/1998

**Abstract:**

*We describe a method for estimating human head pose in a color image that contains enough of information to locate the head silhouette and detect non-trivial color edges of individual facial features. The method works by spotting a human head on an arbitrary background, extracting the head outline, and locating facial features necessary to describe the head orientation in the 3D space. It is robust enough to work with both color and gray-level images featuring quasi-frontal views of a human head under variable lighting conditions.*

# Introduction

The problem of human body parts recognition and detecting their pose in the 3D space has been around for quite a while. A variety of methods has been developed to approach this complex problem. Determining human head pose is just one of many aspects of the mentioned problem [6], [7], [8]. The method described in this paper relies on the idea that the orientation of a human head in the 3D space can be recovered from an image by locating a set of *crucial facial features* within the head silhouette boundaries.

The term *crucial facial features* means that some of the facial features could be a major clue to the head pose recognition. This, of course, depends on the initially assumed view. For instance, if we assume that the person's head is quasi frontal to the camera, we will try to look for such features as eyes and mouth. If, however, the assumption is that the person is looking somewhere to the side, we could be looking for one of the eyes, the mouth and an ear. We shall discuss an algorithm for locating facial features in Facial Feature Candidates Detection section.

The head outline has to be determined before anything else is detected or computed. Our method relies on an existence of a preprocessor routine that segments out the head silhouette for us. Various techniques are available to carry out this kind of work. One approach could be to use the fact that a person cannot sit still for very long; then it is possible to use motion detection techniques to spot an approximate location of the head. Another feasible approach is to use the skin color to segment out a patch corresponding most likely to a face [3], [9]. Any particular implementation of head silhouette detector is beyond the scope of this paper.

# Facial Feature Candidates Detection

Once the head silhouette is obtained, the search for crucial facial features narrows down considerably because then we would know their approximate size and location. It has been noticed that edges can be one of the most informative clues for detecting object's specifics. Then it only makes sense to build an edge map of the head silhouette interior and use this edge information to extract the candidates for crucial facial features.

To determine possible locations of an individual feature, one could build a feature's *location likelihood map* showing how likely a feature is to appear at a particular position on the image. This procedure will sweep the head outline interior with a feature specific mask determining at each location the number points corresponding to an edge versus interior (non-edge) points. Each edge point will contribute to the sub-histogram of edge's normal angles versus brightness. The algorithm will then compare the computed histogram with the model to tell the likelihood of the feature to be at a particular location.

To reduce our analysis to just a few most possible locations, it will be necessary then to point out the best peaks in the location likelihood map. The *non-maxima suppression* procedure will do exactly that. It will use the feature's specific mask size to suppress false peaks around each most probable one.

## Extracting Edges

The edge detection procedure runs in the two main stages. First, the interior of the head silhouette is de-blurred to sharpen potential edges. Second, the edge points are found and the edge's local normal directions are computed.

The de-blurring algorithm uses so called *min-max image sharpening* technique. The procedure processes each pixel in the following manner:

- inspect the 8-neighborhood of it
- find its brightest and the darkest neighbor
- decide which one the current pixel is closer to (by brightness)
- assign the current pixel the value of the closest extreme

Thus at a cost of inexpensive preprocessing, the resulting image will have much sharper edges than the original one.

If the processed image is a color one, it makes sense to detect edges in each of the three color-bands (e.g. detect red, green and blue edges). For a gray scale image, there will be only one band to consider. The edge detector proceeds by inspecting the lower half-8-neighborhood of each pixel. If the difference between any two pixels (in any color band) exceeds certain threshold (specific to the color band), both points are collected to the edge point pool. The resulting collection will contain two-sided color edges from all the three color-bands.

At each edge point, we also compute the *edge's local normal direction*, an approximate direction of the edge's curve normal vector. The edge's local normal provides us with the information about the edge's orientation at a particular edge point. The computation is done by averaging the contrast directions over the point's eight neighbors. The edge's local orientation information is naturally associated and stored with the corresponding edge points from the edge point pool.

## Building Feature's Location Likelihood Map

To spot most probable locations of a crucial facial feature, the method builds a *feature's location likelihood map*. This map reveals how likely a feature is to be located at a given image point within the head silhouette. To build such a map, the method scans its search area with a *feature's mask* and collects data for the *feature's signature histogram* associated with every point of the search area. Then it determines how far the collected signature histogram is from the *model signature histogram* by applying an appropriate histogram distance measure. This computed distance value becomes an entry of the feature's location likelihood map at the image point just scanned.

The *feature's mask* is a set of coordinates of image points that cover the feature along with some small neighborhood around it. The coordinates could be given relative to any point that seem fit. One of the convenient choices could be either center of the mask or one of its enclosing rectangle's corners. This method assumes that the mask is a convex set but, with some caution, this requirement could be dropped.

At each location of the search area, the feature's mask segments out a set of image points to collect the *feature's signature histogram*. For that histogram, the method specifies to have two major bins: the *edge point bin* and the *non-edge point bin*. The *non-edge point bin* contains a count of non-edge points found under the feature's mask. The *edge point bin* unravels to be a two-dimensional sub-histogram of edge's local normal angle versus brightness at an edge point. A histogram built this way is expected to carry enough information to capture individuality of a facial feature, thus the name.

The *model signature histogram* should be pre-computed for each feature's mask by collecting feature's signatures from a set of training images. The average value of the training histograms will produce the *model signature histogram* that can be used for experiments outside the training set.

Finally, when comparing the histograms, a distance measure is needed. One obvious choice could be an $L_1$ norm that computes a sum of absolute differences of the corresponding bins. Another good choice could be the Kullback measure given by the formula:

$$D(s:m) = \sum_i s_i \ln \frac{s_i}{m_i}$$

where *s* is the histogram being tested, and *m* is the model histogram. Notice that the entries of both histograms have to be converted to probabilities.

## Finding Best Feature Candidates

The feature's location likelihood map contains enough information to point out the best feature candidates. The method needs to significantly reduce the pool of potential feature candidates leaving only the best peaks for the best constellation search. Notice that the location likelihood map can be considered a surface of two-dimensional discrete argument whose entries are positive. The problem of finding best feature candidates can be considered as a problem of identifying all local minima at a given domain. To solve it, the method employs the *non-maxima suppression* technique.

The idea of non-maxima suppression is to identify the best peaks (local maximums) in a given array of values and suppress all false peaks around the best ones. This is accomplished using slightly modified priority queue [4]. The difference in the queue behavior is that the newly inserted element, as it ascends the heap, can suppress (be suppressed by) other queue elements in its domain that have lower (higher) ranking. Iterated several times, this procedure will significantly reduce the number of non-significant queue elements. The iterations stop when the queue size is not decreasing anymore.

Because the non-maxima suppression procedure finds local maximums and we are interested in local minimums of the location likelihood map, the method inverses the map entries and feeds them into the non-maxima suppression algorithm. The result is the set of local minimums of the original map.

## Best Constellation Search

Having found the most probable locations for each of the crucial features, the method proceeds with evaluating all possible constellations using random graph matching procedure [1]. The constellation that receives the highest rank is considered the best. Assuming the mutual distances between the features are normally distributed, the method evaluates each feature arrangement $c_i$ in the following manner:

- build the mutual distance vector $v_i$
- scale-normalize it: $w_i = \dfrac{v_i}{\lambda_i}$
- find its probability distribution value $p_i = p(v_i) = N(v_i; \lambda_i \overline{L}, \lambda_i^2 \Sigma)$
- rank the constellation as $r_i = p_i \sum_j c_{ij}$

where $c_{ij}$ is the *j*-th feature's peak value

All distribution parameters and methods of scaling factor estimation are given in [1]. Notice that the procedure above is independent of translation, rotation or scale transformations.

# Head Pose Estimation

To address the head pose estimation problem given the best constellation within the head outline we need to become a little more specific about the crucial features one can use. It is essential that the mutual distances between the facial features of choice be conserved. One simple set that can work for a quasi-frontal view is a pair of eyes and the "mouth and nose-tip" area. We shall discuss how to estimate the head pose using a constellation of these three features. One can generalize this approach for some other set of features.

The features mentioned above make up a triangle *MLR* with the following properties. We expect distances $ML = a^2$, $MR = b^2$, $LR = c^2$ to be in the ranges specified in [2].

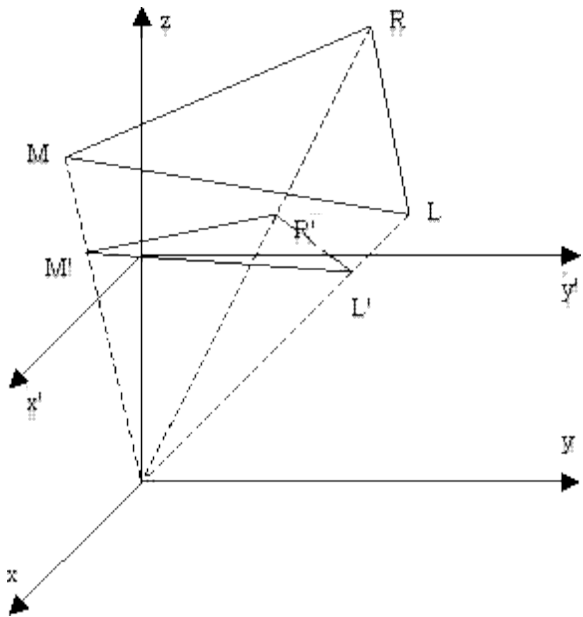

We know everything about triangle *M'L'R'*, the projection of *MLR* onto the image plane. It is our task now to compute the normal of the plane *MLR*.

Stated this way, the problem will always have at least one solution. This follows immediately from the properties of the parallel projection mapping and the triangle properties. Therefore, existence is provided. The solution might not be unique, however. In the most general case, there will be four solutions. This can be shown both geometrically and analytically. We give a justification of this fact later in this section. Let us first discuss how one can choose the solution that gives us the correct head pose estimate.

Given the head silhouette, we can determine an average direction in which the crucial facial features are shifted from their expected frontal positions relative to the center of the silhouette. Call this vector *s*. Out of all normals to MLR found, we should choose the one whose projection is the *most* co-directional with *s*.

Let O'*X'Y'* be our image plane as Figure 1 shows. One then can write the following system of nine equations for nine unknowns:

$$\begin{cases} (X_M - X_L)^2 + (Y_M - Y_L)^2 + (Z_M - Z_L)^2 = a^2 \\ (X_M - X_R)^2 + (Y_M - Y_R)^2 + (Z_M - Z_R)^2 = b^2 \\ (X_R - X_L)^2 + (Y_R - Y_L)^2 + (Z_R - Z_L)^2 = c^2 \\ \dfrac{X_L}{X'_L} = \dfrac{Y_L}{Y'_L} = \dfrac{Z_L}{f} \\ \dfrac{X_R}{X'_R} = \dfrac{Y_R}{Y'_R} = \dfrac{Z_R}{f} \\ \dfrac{X_M}{X'_M} = \dfrac{Y_M}{Y'_M} = \dfrac{Z_M}{f} \end{cases}$$

It should be clear by now that this system shall produce more than one solution. There are also many ways we can solve this system. One could attempt to solve this nonlinear system directly using some parallel computation techniques as it was proposed in [11]. Another approach is to reduce the system to a single polynomial equation of degree four as it has been shown in [10]. The former could be efficient and quite reliable but requires parallel hardware. The latter might seem a little more elegant and does the job utilizing a single processor but involves some tedious algebraic computations and still leaves the programmer with the task of solving the polynomial equation. In any case, the actual implementation would be application specific. Once the system is solved, the triplet *MLR* with the best normal is chosen as described above.

# Conclusions

A method for estimating human head pose from a color image has been developed. The described method has several stages:

- head silhouette estimation
- crucial facial features location
- best constellation search
- estimation of the head pose using facial features location within the head silhouette

The source image was assumed to carry enough information for the algorithm to estimate a person's head outline and to run a color edge detector to locate non-trivial edges of certain facial features. Although the head silhouette detector has not been described, some ways of building one were discussed in the Introduction section. The facial feature location involved a color edge detector whose output was fed into facial feature pattern analyzer that used histogram matching technique to spot most probable locations of the crucial facial features. The pool of the candidate features was processed to find the best constellation. Assuming the mutual distances between facial features are normally distributed, the best constellation search procedure used random graph matching method to find the best feature arrangement. By processing the best constellation position within the head silhouette, the method computed the three rotation angles describing the person's head pose in the 3D space.

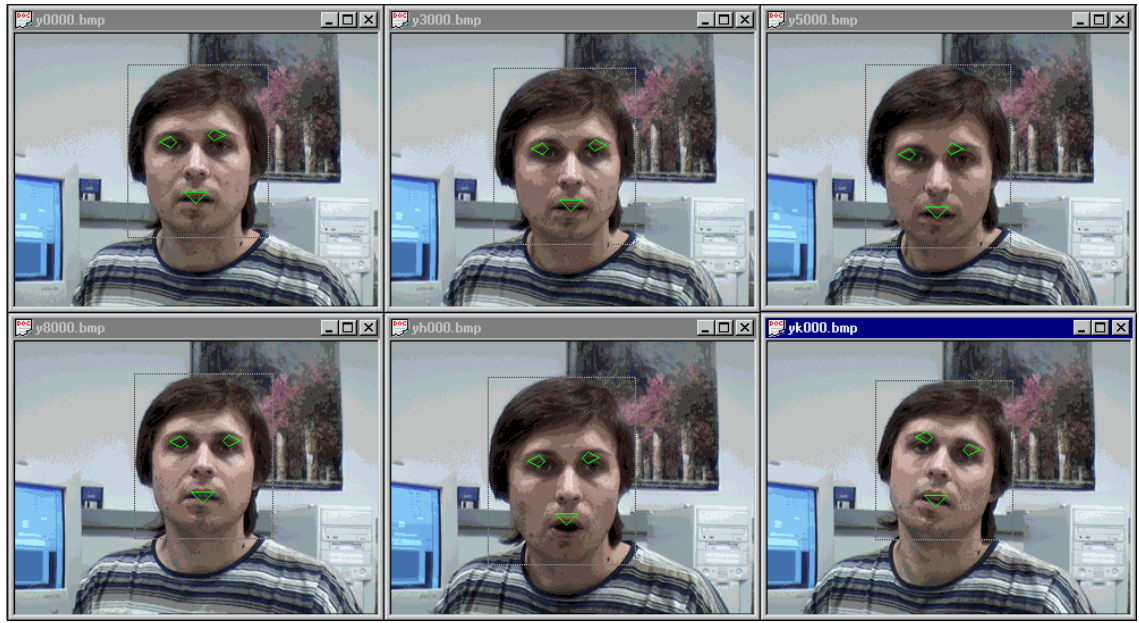

The method has been tried on a series of 320x240 true color images each displaying a single individual in quasi-frontal view. It was shown that the method could reliably estimate a person's head pose, provided the head silhouette was pre-segmented correctly and the crucial facial features were not obscured by anything. The facial feature locator was pre-trained on a set of quasi-frontal view images. The location of facial features was employing separate masks specific to each feature. Parameters for the normal distribution used in the best constellation search procedure were drawn from [1] and [2].

The overall accuracy of the method, of course, depends on how accurately the head silhouette is estimated. The method, however, gave correct enough estimate of the head pose even if the head outline was given approximately. With all its achievements, the described method yet opens some space for further development. It, for instance, could be adapted to be a (close to) real-time head pose tracker. One can also abstract it to estimate an orientation of any object in the 3D space, given it has a set of crucial features the method can identify and rely on.

Notice that presented method spends a lot of time locating the head silhouette and then locating the facial features within it. It does not make any use of the information it recovered from the frame it just processed. If it is assumed that a video sequence will be processed, then the difference between the two consequent frames will be small. Instead of searching for the head silhouette and the facial features every time from scratch, the method could locate them once, and then, track them. With the narrowed search area, the head pose estimation could then be done very fast.

For the method to be adapted to estimate an orientation of an arbitrary object in the 3D space, certain requirements should be met. First, the object should possess a set of identifiable and distinguishable features, or marks. Second, the mutual distances between the features should be conserved. The method should become aware of the object's shape or some aspects of it. Finally, there should be a routine, which can build the object's outline. Having all that, the method should be able to correctly estimate the three cosines necessary to describe the object's orientation in the 3D space.

**Acknowledgements**

I express my thanks to Larry Davis, professor of UMCP, who performed high-level supervision and provided useful detailed feedback at all stages of the project. My special thanks are to David Harwood, associate professor of UMCP, whose assistance in development of the described method was crucial, especially at its early stages.